# Mimicry Without Understanding: The Origins of Decision Bias in Large Language Models

Eldad Yechiam[1*] and Adi Tarabeih [1]

1. Technion – Israel Institute of Technology, Israel, Haifa

*Address for correspondence: Eldad Yechiam, Faculty of Data and Decision Sciences, Technion - Israel Institute of Technology, Haifa 3200003, Israel. Email: yeldad@technion.ac.il

**Mimicry Without Understanding: The Origins of Decision Bias in Large Language Models**

# Abstract

Large Language models (LLMs) were found to be susceptible to a host of social, affective, and cognitive biases. We examined two mechanisms through which such biases can be generated even when human preferences (in the training data) are not biased or when they are correctly categorized as being biased. The first is faulty mimicry of preferences based on human behavior: this involves LLMs inferring human preferences even when behaviors are logically unrelated to preferences. The second is mimicry of explicitly biased human behaviors. In four studies focusing on economic biases, we find that ChatGPT-4o and Qwen exhibited social proof biases even when prompted with reports of human behaviors that were clearly non-indicative of individuals' actual preferences. LLMs also displayed loss aversion when it was explicitly described as a bias. Indeed, when prompted with detailed scientific reports, the extent of the bias (i.e., loss aversion) in the scientific report predicted LLMs' own subsequent bias. Scientific papers of biases can thus become self-fulfilling prophecies, at least when it comes to LLMs' responses. The current study goes beyond fleshing out LLM biases and sheds light on the underlying component processes.

# Introduction

Large language models (LLMs) generate responses through probabilistic modeling of linguistic patterns learned from large-scale datasets. Initially, it has been argued that if these linguistic patterns include the laws of rationality (i.e., probabilistic reasoning, Bayesian updating, etc.), LLMs should display economic rationality far exceeding that of humans [1]. Yet it was soon found out that in many domains LLMs exhibit considerable human-like biases. In the social domain these include, for instance, self-other attribution gaps [2], ingroup favoritism and outgroup disparagement [3], gender predispositions [4,5], and even political biases (reportedly favoring Democrats in the US, Lula in Brazil, and the Labour Party in the UK [6]). In non-social judgments LLMs were shown to rely on completely irrelevant information to make judgments (i.e., anchoring [7-10]), judge the probability of events based on anecdotal information (availability heuristic [8]), estimate the probability of two co-occurring events as higher than the likelihood of either of them alone (i.e., representativeness heuristic [8]), and dichotomize statistical evidence [11]. Even in economic decisions between lotteries defined by their probabilities and outcomes, where expected-value (EV) calculations are relatively easy, LLMs exhibit consistent biases, such as overweighting small-probability events in decisions from descriptions and underweighting these events in decisions from experience, as humans do [12-13]. What leads to these biases? The present paper suggests two processes of bias generation that go beyond sheer mimicry of humans' biased preferences and lead to extreme sensitivity to documented human behaviors that form the corpus of the LLM's training data.

The first such process is faulty interpretation of human preferences from human behavior. Previously, it has been proposed that LLM biases are the product of mimicking the behavior of people described in the corpus of training data, while completely ignoring the rules of rationality [14]. In other words: LLM see, LLM do. We argue that while this may indeed be the case, such mimicry emerges even in cases where human behavior is not indicative of relevant preferences. For example, if I choose two US dollars over one yen, this doesn't actually bear on my weighing of same-number dollars and yen, besides setting a weak constraint (2 yen < 1 USD); yet, we posit that LLMs consider the sheer behavior of choosing as indicative of preference, in this case inferring that USD are preferred over yen, independently of logical considerations.

The second process that arguably facilitates biases in LLMs is that mimicry of human behavior persists even when biases are not implicit but are explicitly described as biases, namely when there is an explicit disclaimer that the relevant behavior exemplifies bias. We suggest that this process potentially leads to a bizarre phenomenon whereby scientific empirical studies of biases, which clearly describe human participants' behavior as being biased, paradoxically facilitate the biases exhibited by LLMs. Moreover, given LLMs' sensitivity to anecdotal information noted above, such facilitation of biases may emerge even when human data is based on a rather small sample, namely when individuals' behavior does not reliably represent the population.

Importantly, the two processes described above may not operate unconditionally. Their influence could be modulated by the extent to which mimicry conflicts with rational decision laws. One such universal law is expected-value maximization, namely picking the option that provides the best outcome based on its expected value (probabilities multiplied by the outcomes they yield). Under this account, the tendency to mimic human biases would be curbed, at least partially, when doing so reduces the LLM's expected outcomes: For instance, if a person is biased to give losses more weight than gains, then they are said to be loss averse [15]. Given that an LLM is informed that humans behave in a loss averse manner, then under the former account (LLM see, LLM do) it would display loss aversion irrespective of the consequences. Alternatively, under the notion of expected-value minded mimicry, the LLM may display loss aversion when there are no downsides to being loss averse (in terms of expected value), but *less so* when exhibiting this bias is considerably disadvantageous.

In four studies, we evaluated the emergence of the two faulty processes of mimicry of human behavior, and their boundary conditions. Our first study used a setting where LLMs are exposed to subjective human preferences between currencies that are not consistent with their objective value. Specifically, we tested whether a safe (fixed amount) monetary option is judged more attractive when denominated in people's preferred currency, even though the relevant currencies are objectively equivalent (1:1 exchange rate). In other words, we evaluated whether LLMs would display a social proof bias: Assuming that something is good or correct simply because others endorse it [16-17].

Social proof bias is not necessarily irrational or illogical because LLMs may assume there are some unknown reasons to prefer one of the currencies. In our second study we therefore considered a setting where people's behavior is clearly non-indicative of their preferences (similar to the 2 USD > 1 yen example above). We examined whether LLMs would show a social proof bias in this setting as well, which would be completely at odds with logical inference.

Our next studies evaluated the process of mimicry where the adopted human behavior is explicitly described as a bias. We focused on loss aversion, the tendency to overweight losses compared to corresponding (i.e., symmetric) gains [15]. Loss aversion is considered a bias because from a strictly objective point of view, removing a certain amount of money and adding the same amount should have the exact same cost/benefit, especially for small stakes that do not have additional externalities. In recent empirical studies it has been disputed whether people are loss averse on average [18-23]. We conjectured that LLMs' loss aversion would mimic that of humans, even if the relevant group of humans is very small. In Study 3, we evaluated if LLM agents prompted with experimental results that people are either loss averse or gain seeking (i.e., the opposite of loss averse), adopt the same bias given various costs, and when presented with the behavior of a large or a small sample. A similar degree of mimicry of a small and large sample attests to shallow copying since it does not take into consideration that average behavioral trends, particularly in small samples, may be driven by random noise.

In our final study we examined the effect of prompting LLMs with texts from scientific papers that explicitly describe loss aversion as a bias, and that report either strong or weak loss aversion, and testing the models' subsequent loss-aversion bias. Importantly, this enabled testing whether an LLM would copy the human bias (loss aversion) despite the clear indication in the scientific paper that being loss-averse is a "disproportionate" (p. 585) tendency [22] which reduces expected value in the studied decision tasks.

Our studies focused on ChatGPT-4o [24] and Qwen 2.5-72B-Instruct [25]. ChatGPT is one of the most culturally prominent LLMs available today (with over 800 million weekly active users globally by 2025 [26]). Much of the research demonstrating biases in LLMs, reviewed above, was conducted on ChatGPT. We also replicated our initial studies with an open source LLM, Qwen [25].

# Materials and methods

## Study 1

To study LLMs' basic mimicry processes, we presented a scenario where two currencies, Tenits and Tanas, have the exact same objective value yet humans subjectively prefer one of the currencies. In one condition people were described as preferring Tenits (Tenits>Tanas) while in the other they were described as preferring Tanas (Tanas>Tenits). We tested whether people's sheer subjective preference for one of the currencies evolves into a similar model-preference when an LLM is exposed to it. We also examined if the adopted subjective preferences would persist when they lead to reduced expected value.

### Data

We conducted 400 independent sessions via OpenAI's API of ChatGPT-4o and Hugging Face Inference API of Qwen version Qwen2.5-72B-Instruct, with 100 agents for each of the two study conditions in each LLM. These two respective versions were the most advanced versions at the time of the study enabling inputs from large text files (up to 128,000 tokens). Both LLMs' temperature parameters were kept at the default level (0.7). Each session was considered to represent a separate entity (or agent) randomly sampled from the distribution of model-generated instances, and analogous to an experimental participant [27]. A Python script managed all interactions. No conversational history or state was carried over between runs, ensuring that the model processed the full prompt anew each time.

### Procedure

The LLMs were informed as follows: "Tenit and Tanas are two objectively equal currencies of the same country (1:1 exchange rate) but people tend to treat them differently. Specifically, people were found to prefer 11 Tenits over 12 Tanas; people were found to prefer 26 Tenits over 28 Tanas; people were found to prefer 109 Tenits over 111 Tanas; people were found to prefer 407 Tenits over 413 Tanas." This version was labeled as the Tenits>Tanas condition. In the alternative Tanas>Tenits condition the reverse preferences were prompted (e.g. "people were found to prefer 26 Tanas over 28 Tenits").

Next, LLMs were asked about their own preferences between a risky lottery (producing a 50% chance to win 10 Tenits and 50% to win 10 Tanas), and each of six safe outcomes as described in Box 1. These six items (where LLMs select between the lottery and a safe option) enable testing the preference for Tenits vs. Tanas when the safe option amounts to 100% of the lottery's expected value, 80% of its expected value, or 50% of the lottery's expected value, respectively. Given that the safe option involves the currency of choice, the smaller the safe option's amount the more LLMs stand to lose as a result of their subjective preference for the currency (compared to the lottery where it is obtained with only 50% chance). For example, if the safe option is 5 Tanas, liking Tanas and preferring this option leads to a 50% reduction in expected value (compared to the lottery's EV of 10). The possibility that this safe choice is due to risk aversion is controlled for by the presence of a 5 Tenits safe option. The effect of subjective preference is thus measured by comparing the case where the safe option is in the preferred currency to where it is not.

In the Tanas>Tenits condition, the questions were identical except that the currency of choice (Tanas) was described first in the lottery. The study was run in an open format where the LLMs were not initially required to produce a structured answer. For the Qwen model, this often resulted in non-convergent or verbose outputs where the model failed to commit to a clear decision. Therefore, an explicit instruction was added to the prompt, directing the agents to provide only a final answer (see details in the supplementary section).

### Analysis

Our main analysis was a logistic Generalized Mixed Model (GMM) with subjective preferences for the safe outcome currency (preferred or not preferred) and safe option amount (5, 8, 10) as within-subject factors, and with the label of the un/preferred option (Tanas>Tenits vs. Tanas<Tenits) and model (ChatGPT vs. Qwen) as between-subject factors. A random-effect intercept was included for each agent. The model with the four-factor interaction (preference, safe option amount, label, and LLM) did not converge and therefore this particular interaction was removed.

## Study 2

The design of this study was similar to that of Study 1 except that choice-related information was clearly non-indicative regarding humans' preferences.

### Data

As in Study 1, we conducted 400 independent sessions via OpenAI's API of ChatGPT-4o and Hugging Face Inference API of Qwen version Qwen2.5-72B-Instruct, with 100 agents for each study condition in each LLM.

### Procedure

The LLMs were informed as follows: "Tenit and Tanas are two objectively equal currencies of the same country (1:1 exchange rate) but people tend to treat them differently. Specifically, people were found to prefer 12 Tenits over 11 Tanas; people were found to prefer 28 Tenits over 26 Tanas; people were found to prefer 111 Tenits over 109 Tanas; people were found to prefer 413 Tenits over 407 Tanas." Notice that this version, labeled as the Tenits>Tanas condition, actually does not imply that people prefer Tenits over Tanas but merely sets an upper bound on the valuation of Tanas (e.g., the first sentence implies Tanas are not valued over 12/11 Tenits). In the alternative Tanas>Tenits condition the reverse preferences were prompted (e.g. "people were found to prefer 28 Tanas over 26 Tenits").

Otherwise, the procedure was identical to that of Study 1 (see details in the supplementary section). For GPT-4o, in 3.5% of the cases LLMs did not select any of the options. In order to avoid missing records, we coded such responses as intermediate since the algorithm did not show a preference (Safe = 1, Lottery = 0, Intermediate = 0.5). Removing these records replicates all of the findings reported below.

### Analysis

The analysis was the same as in Study 1 except that because of the no-preference responses the dependent variable was rank-based, using proportional odds (i.e., a Cumulative Link Model).

## Study 3

We presented LLMs with (fictitious) experimental results that indicated either considerable loss aversion or gain seeking on the part of human participants. An additional feature of the study was that the degree of departure from expected-value maximization as a result of being biased (due to loss aversion or gain seeking) varied in different items. Finally, we compared LLMs' mimicry of human biases in response to large-sample experiments versus small and relatively unreliable experimental samples.

## Data

We conducted 800 independent sessions via OpenAI's API of ChatGPT-4o and Hugging Face Inference API of Qwen version Qwen2.5-72B-Instruct, with 100 agents for each of the two (gain seeking vs. loss aversion) by two (small vs. large sample) between-subject conditions, in each LLM. The script was run as in Studies 1 and 2.

## Procedure

LLMs were informed as follows: "In an experiment, a fair coin was tossed (50% chance of it turning up tails, and 50% heads). People were offered an opportunity to enter a lottery based on the toss of the fair coin. If the coin turns up heads, they lose some money; if it turns tails, they win some money. This experiment included questions about five lotteries. For each lottery, participants could accept the lottery (winning or losing according to the coin toss) or reject the lottery (and get zero instead)." Next, the six lotteries were described, as indicated in Box 2.

Then, agents in the loss-aversion condition were provided with the following text: "In this very large (10,583 people) / very small (43 people) experiment, it was found that most people (about 60%) accepted Lottery A, half of the people (about 50%) accepted Lottery B, most people (about 55%) rejected Lottery C, most people (about 60%) rejected Lottery D, most people (75%) rejected Lottery E, and most people (about 100%) rejected Lottery F." Note that the size of the experiment was different in the two between-subject conditions that contrasted a large (n = 10,583) and a small (n = 43) sample. Note also that the modal human participant described here was loss averse since most participants rejected Lottery D where losses are slightly smaller than gains (-$5, $6). By contrast, in the gain-seeking condition, the following alternative text was presented: "In this very large (10,583 people) / very small (43 people) experiment, it was found that most people (about 100%) accepted Lottery A, most people (about 90%) accepted Lottery B, most people (about 85%) accepted Lottery C, most people (about 80%) accepted Lottery D, most people (65%) accepted Lottery E, and most people (60%) rejected Lottery F." Here, most people accepted the lottery with symmetric gains and losses, and hence the modal human behavior can be described as gain seeking. Finally, LLM agents were presented again with the six lotteries, and following each were asked to indicate whether they accept or reject it (see Supplementary section for full prompts).

## Analysis

The effect of study condition was examined with an ordinal-logistic regression analysis for the total number of "accept" decisions as a (reverse) behavioral indicator of loss aversion. The regression model involved three between-subject factors: The loss aversion vs. gain seeking condition based on the experiment described to the model, the experiment's sample size (small vs. large), and the model (ChatGPT vs. Qwen). We also extracted the models' loss aversion parameter using prospect theory [15]. This was done in a simple form (as in [28,29]): The loss aversion coefficient was estimated for each individual model agent by computing a) the constant gain (i.e., $6) divided by the smallest loss for which the lottery was not accepted and b) the constant

gain divided by the largest loss that was accepted, and taking the mean of the two resulting ratios. Thus, for example, if the smallest loss for which the lottery was not accepted was \$5 then the actual threshold for accepting is anywhere between \$4 and \$5. Hence, $\lambda$ would be estimated as (\$6/\$5 + \$6/\$4)/2 = 1.35. We also ran a more orthodox estimation of $\lambda$, described in the supplementary section.

## Study 4

In three conditions we asked LLMs to base their response on the abstract of each of three meta-analyses involving loss aversion: a) Brown et al. [22] who found an estimated mean $\lambda$ of 1.96 for 150 studies, implying very strong loss aversion (a loss is weighed almost twice as much as a corresponding gain), b) Walasek, Mullett, and Stewart [30] who found a lower $\lambda$ estimate of 1.31 for 17 studies, and c) Yechiam and Zeif [23] who re-analyzed Brown et al.'s [22] data and found a $\lambda$ of only 1.07 for lotteries with equal sized gains and losses and no ordering, thus showing very weak (and non-significant) loss aversion. Additionally, we compared these three conditions to a control condition where the LLM did not base its response on any scientific paper. We expected the model to display loss aversion coherently with the prompted scientific paper's loss aversion.

### Data

We conducted 800 independent sessions via OpenAI's API of ChatGPT-4o and Hugging Face Inference API of Qwen version Qwen2.5-72B-Instruct, 100 in each of the four study conditions (three abstracts and a control condition). The script was run as in the previous studies.

## Procedure

The LLM was informed as follows: "Answer the following questions using *only* the information provided below." It was next presented with the title and abstract of either Brown et al. [22], Walasek et al. [30], Yechiam and Zeif [23], or no text (control condition). In the control condition the LLM was asked to simply answer the questions. Next, the LLM was presented with the six items shown in Box 2, all of which involve the decision of whether to accept or reject a lottery producing gains and losses. In the current study, three additional items were added where the loss was increased to $8, $9, and $10, enabling a more refined quantification of gain seeking.

## Analysis

An ordinal-logistic regression analysis was used for the total number of "accept" decisions. The regression model involved two between-subject factors: The four study conditions (three meta-analyses and a control condition) and the model (ChatGPT vs. Qwen). The loss aversion parameter was estimated as in Study 3.

# Study 4b

In this study we prompted ChatGPT with the entirety of the three meta-analyses concerning loss aversion. For Qwen this was not possible due to the length of the text and the tendency of the algorithm to degrade its performance given such length [25].

## Data

We conducted 400 independent sessions via OpenAI's API of ChatGPT-4o, 100 in each of the four study conditions. The script was run as in the previous studies.

### Procedure and analysis

The procedure was as in Study 4a except that the scientific papers' full text was included in the prompt. The papers were presented without author details, the reference section, and their figures. The analysis was as in Study 4a.

# Results

The studies' complete data is available at https://osf.io/xj7ba/overview?view_only=28d95db8ccb04fcc8d2743e6f0978388 .

## Study 1

Figure 1 left panels present the LLMs' safe-option choices when it was denominated in the currency subjectively preferred by humans or not, and under different values of the safe option. The figure pools across the different labels of the preferred currency (Tanas vs. Tenits) for brevity. As can be seen, in all conditions there was a strong preference for the currency preferred by humans. Additionally, this human-mimicked preference diminished as the cost of making it increased (most markedly in the S = 5 condition). The GMM was significant, $\chi^2(1) = 1{,}340.97$, $p < .001$, and accounted for 77.6% of the variance. The model indicated a significant effect of the human (subjective) preference for a particular currency, $\chi^2(1) = 116.61$,

$p < .001$, Cohen's $d = 0.61$, and also a significant main effect of the safe option's amount in line with its expected value, $\chi^2(1) = 285.12$, $p < .001$, $d = 0.56$.

Importantly, there was also a significant interaction between the subjectively preferred currency and the safe amount, $\chi^2(1) = 23.93$, $p < .001$, denoting a decrease in mimicry as the expected value of adopting the human preferences declined.

There were no significant differences between models (ChatGPT vs. Qwen) in the mimicry tendency, as described in Supplementary Table 1.

To what extent is the mimicry tendency reduced when it implied a considerable cost? In the S = 10 condition, where no cost is involved for adopting the human tendency, 37.5% of the LLM agents (pooled across ChatGPT and Qwen) based their response on human subjective preferences and selected the safe option when it was in the preferred currency (vs. not). By contrast, in the S = 5 condition, where having this preference diminishes expected value by half, only 17.0% of the agents selected the safe option when it was preferred by humans (vs. not), thus showing a small, but still significant bias ($\chi^2(1) = 81.24$, $p < .001$). This indicates that LLMs' propensity to mimic based on reported preferences decreased but was not eliminated when it reduces expected value.

## Study 2

Figure 1 right panels present the LLMs' safe-option choices when it was denominated in the currency chosen by humans, even though this was irrelevant to humans' bona fide preferences between the two currencies. As can be seen, in all conditions LLMs strongly preferred the currency chosen by humans, as in Study 1. The GMM was significant, $\chi^2(1) = 1{,}195.44$, $p < .001$, and accounted for 69.7% of the variance. The model showed a significant effect of implied human preference for a particular currency, $\chi^2(1) = 227.82$, $p < .001$, $d = 0.72$, even though this implication is logically incorrect. There was also a significant main effect of the safe option's amount in line with its expected value, $\chi^2(1) = 331.49$, $p < .001$, $d = 0.52$, and a significant interaction between the subjectively preferred currency and the safe amount, $\chi^2(1) = 40.66$, $p < .001$. The latter effect denotes the diminishment of the mimicry tendency as the cost of making it increased (most strongly in the S = 5 condition). Comparing the degree of mimicry in the current study to that of Study 1 where mimicry was based on factual preferences, the results indicate that the magnitude of mimicry is similar in both studies (0.72 compared to 0.82, respectively).

Did the LLMs infer that people preferred one of the two currencies over the other? A review of ChatGPT-4o's verbal responses indicates that such inferences were made in every single query (100% of the agents), and that the preference was based on human choices; namely, when humans chose "28 Tenits over 26 Tanas" this was interpreted as a human preference for Tenits over Tanas.

## Study 3

Figure 2 presents the rate of "accept" responses for each of the items in the four conditions and for each loss involved. As can be seen, in the loss-aversion condition, ChatGPT and Qwen agents mimicked the human sample and tended to accept fewer lotteries involving potential losses, irrespective of the experiment's sample size. For ChatGPT this was, interestingly, most prominent when the loss was only slightly smaller than the gain (loss of $5 compared to gain of $6) whereas for Qwen the largest difference plateaued between losses of $4 and $6. A regression analysis indicated that the saturated model did not converge when adding all interactions, and hence we ran a model including only the interaction between study conditions, $\chi^2$ (4) = 719, $p < .001$, $r^2_{MCF} = 0.35$ (a linear model with all interactions is described in Supplementary Table 2). The results indicated a significant effect of the human preference (loss aversion vs. gain seeking) on the rate of "accept" decisions, B =

17.92, SE = 0.001, Z = 24,502, $p < .001$, along with a nonsignificant effect of experimental sample size (small vs. large), B = 0.13, SE = 0.19, Z = 0.68, p = .50; and importantly, no significant interaction between the human preference and the sample size, B = -0.09, SE = 0.28, Z = 0.32, p = .75. The latter finding indicates that mimicry is independent of the number of target individuals and whether they reliably represent the human population. In addition, there were some differences between the two LLMs detailed in the supplementary section.

We next calculated the loss aversion parameter $\lambda$ for each individual agent. For ChatGPT, in the loss-aversion condition the average $\lambda$ parameter equaled 1.257/ 1.218 for the large/small sample size conditions, respectively. Yet, in the gain-seeking condition the average $\lambda$ equaled 1.072 in both sample sizes. For Qwen agents, the loss-aversion condition produced an estimated mean $\lambda$ of 1.742 for both the large and small sample sizes, while the gain-seeking condition led to a mean $\lambda$ of 0.940 or 0.929, respectively. Thus, the prompted behavior of human participants in the experiment toggled the LLMs' loss-aversion tendency, almost irrespective of sample size.

As previously, we examined whether mimicry diminishes as the bias reduces expected value. This was not quite the case in the present study. Instead, the LLMs seemed to be most sensitive to prior human preferences at the point of indifference between options: For ChatGPT the effect of human preferences (loss aversion vs. gain seeking) was more pronounced at -$5, +$6, where being loss averse implies a small expected-value reduction; than at -$6, +$6: B = 1.46, SE = 0.34, t = 4.30, $p < .001$, where loss aversion does not affect one's expected value. Qwen showed the same pattern though the disparities were tiny and not statistically significant. Potentially, this was due to a slight predisposition to loss aversion prior to the information that we presented which is also consistent with the larger influence of the loss-aversion condition than the gain-seeking condition.

## Study 4a

Figure 3 top and middle panels present the results of the study. As can be seen, the prompted abstract affected the model's accept/reject decisions. For both LLMs the largest number of "accept" decisions emerged when basing the answer on Yechiam and Zeif [23], followed by Walasek et al. [30], and finally Brown et al. [22]. The control condition was close to the results based on Yechiam and Zeif [23] (slightly more loss averse for ChatGPT, gain seeking for Qwen). The regression model was significant, $\chi^2$ (7) = 1,043 , p < .001, $r^2_{MCF}$ = 0.58, with the difference between conditions being significant for all three contrasts, Z > 6.69, p < .001. In addition, there was a significant effect of the model (ChatGPT vs. Qwen), Z = 11.59, p < .001; and a significant interaction between the study condition and the model, Z > 6.16, p < .001 for all contrasts. For ChatGPT, the difference between conditions was very large, with 105% more lottery acceptance decisions based on Yechiam and Zeif [23] than on Brown et al. [22] For Qwen the difference was smaller, with only 9% difference between studies.

Figure 3 right panel shows the mean $\lambda$ parameters estimated from LLMs' responses as a function of the average $\lambda$ reported in the scientific paper's abstract. As can be seen, there was a monotonic relationship between how human study participants were reported to behave and how the LLM decided: $\lambda$ was highest (ChatGPT: 1.785, Qwen: 1.170) in response to Brown et al.'s [22] report of strong loss aversion, and lowest (ChatGPT: 1.043, Qwen: 1.047) in response to Yechiam and Zeif's [23] report of no significant loss aversion, thus being consistent with the degree of loss aversion reported in the human studies.

## Study 4b

Figure 3 bottom panel presents the results. As can be seen, the findings replicate those of Study 4a: The largest number of "accept" decisions was based on Yechiam and Zeif [23], followed by the control condition, Walasek et al. [30], and finally Brown et al. [22]. The difference between conditions was significant, $\chi^2$ (3) = 313, $p < .001$, $r^2_{MCF} = 0.33$, as were all contrasts between conditions: $Z > 4.09$, $p < .001$. Moreover, the extent of the difference was very large, with 51% more lottery acceptance decisions based on Yechiam and Zeif's [23] paper than on Brown et al.'s [22] paper. ChatGPT's mean loss aversion parameter was aligned with the loss aversion reported in these papers for human participants, being highest (1.519) when prompted with Brown et al. [22] and lowest (1.082) when prompted with Yechiam and Zeif's [23] paper.

# Discussion

Why are large language models biased in so many domains? It has been generally argued that LLMs simply mimic what people are reported to do [14]. We expanded this notion and evaluated the effect of two additional component processes: One is illogical preference-deduction from people's behavior; the other is mimicking preferences despite being informed that these are examples of biased behavior. To evaluate these component processes, we studied LLMs' decisions in contexts progressively similar to how people's behavior is described in scientific reports: A description of the population's behavior (Study 1 and 2), a description of an experimental sample in a fictitious study (Study 3), and actual scientific reports (Study 4a,b). In addition, we examined to what extent these processes diminish when exhibiting bias is directly disadvantageous in terms of expected value.

Our first study showed that even though LLMs were informed that two currencies are objectively equal, they nevertheless assigned more value to the currency preferred by humans. This social proof bias diminished when the human-mimicked preference implied losing money, namely upon making a decision that goes against expected value. Still, even when expected value was undermined, the adopted bias did not completely disappear. Study 2 went further and examined whether in the same setting LLMs would prefer the currency chosen by humans even though their choice bears no logical implication for their preference (since the chosen currency type was confounded by its size). Interestingly, we found a similar magnitude of mimicry of human behavioral choices in this study. The LLMs preferred the chosen currency despite no logical reasons to do so, following human behavior at the surface level without correctly inferring preferences.

Study 3 was designed to evaluate whether LLMs would mimic a clearly biased behavioral tendency of experimental participants. As in Study 1, the bias reported in the human experiment affected that exhibited by the LLM. If humans were described as loss averse, LLMs showed more loss aversion than if humans were described as gain seeking. Importantly, though, LLMs' tendency to mimic human biases was not sensitive to the size of the sample in the human experiment presented to them: LLMs' loss aversion bias was about equal when the relevant report of human behavior was based on a very large sample (> 10,000) or a very small sample (<50). In the latter case, individuals' behavior may not reliably represent that of the population: Indeed, for this to occur the effect size of loss aversion would have to be exceptionally large. Hence, this again reflects a process of copying human behavior without applying logical – in this case, statistical – inferences.

Also, as found in Study 1 and 2, the mimicry tendency was moderated by the expected value of the biased behavior. Yet, the adoption of human biases was not highest at the objective indifference point, namely where being loss averse did not affect expected value, but rather at a point where the loss was slightly smaller than the gain (which is the indifference point given weak loss aversion). This suggests a refinement on the expected-value moderation hypothesis: Mimicry seems to be largest when the bias leads to the smallest deviation from the model's expectancy (which may be biased) rather than expected value: For an already slightly biased model, the point of indifference is based on the existing bias (the fact that LLMs were slightly biased towards loss aversion was further established in Study 3 control condition). This highlights the need to consider the LLM's own reference point when quantifying economic indifference [31].

Finally, Study 4 used actual scientific abstracts (Study 4a) and papers (Study 4b) to test whether despite the scientific portrayal of the respective human behavior as a bias, LLMs would still mimic the biased human behavior. The results showed that both ChatGPT's and Qwen's loss aversion was significantly and substantially higher as a function of the degree of loss aversion reported in the scientific paper. For example, without relying on any scientific paper ChatGPT had only a very minor loss-aversion bias: Its loss aversion parameter was 1.09, implying about 10% greater weighting of losses than gains. Conversely, after being exposed to Brown et al.'s [22] abstract reporting strong loss aversion, ChatGPT's loss aversion parameter was 1.79, implying that it gave losses about 80% more weight. Even when exposed to the complete report of Brown et al. [22], ChatGPT's loss aversion parameter was 1.52, denoting about 50% greater weighting of losses. Much weaker loss aversion was exhibited when the LLM based its response on papers reporting moderate or no loss aversion. Thus, paradoxically, the human studies that highlighted the strongest deviation from rationality produced LLM behaviors that were the least consistent with rationality, leading to a seeming self-fulfilling prophecy.

Our findings thus indicate extensive mimicry of human biased behavior by LLMs which is surface level and illogical. It occurs when individuals' choice behavior is not representative of the population preferences and even when it is not indicative of their own preferences, and it emerges even where choice behaviors are explicitly labelled as being biased in scientific papers.

Follow-up studies should address whether and to what extent this type of mimicry is susceptible to debiasing. For example, one avenue of debiasing in human studies is requesting individuals to simultaneously consider a list of similar decisions varying in their relevant parameters. This is known as "joint evaluation" or "consider-the-alternative" [32,33]. We implemented this type of debiasing by re-running Study 2 with ChatGPT agents who simultaneously provide accept/reject responses for all lotteries A-F, instead of separately after each lottery (see supplementary section for details). The results, summarized in the supplementary section, showed that the difference between conditions in the LLM's responses (as a result of experimental participants' behavior), became evident only when loss aversion did not conflict with expected value calculations, namely when losses and gains were equal. Thus, while popular and scientific documentation of human biases can exacerbate LLM biases, proper design of prompts and algorithms inspired by the science of behavioral economics may substantially reduce this tendency.

# References

1. Chen Y, Liu TX, Shan Y, Zhong S. The emergence of economic rationality of GPT. P Natl Acad Sci USA. 2023;120:e2316205120.
2. Suzgun M, Gur T, Bianchi F, Ho DE, Icard T, Jurafsky D, et al. Language models cannot reliably distinguish belief from knowledge and fact. Nat Mach Intell. 2025;7:1780-1790.
3. Hu T, Kyrychenko Y, Rathje S, Collier N, van der Linden S, Roozenbeek J. Generative language models exhibit social identity biases. Nat Comput Sci. 2025;5:65-75.
4. Etgar S, Oestreicher-Singer G, Yahav I. Implicit bias in LLMs: Bias in financial advice based on implied gender. 2024. Available at http://dx.doi.org/10.2139/ssrn.4880335
5. Nadeem M, Sohail SS, Cambria E, Afreen S. South Asian biases in language and vision models. Nat Mach Intell. 2025;7:1775-1777.
6. Motoki F, Pinho Neto V, Rodrigues V. More human than human: Measuring ChatGPT political bias. Public Choice. 2024;198:3-23.
7. Castello M, Pantana G, Torre I. Examining cognitive biases in ChatGPT 3.5 and ChatGPT 4 through human evaluation and linguistic comparison. In Proceedings of the 16th Conference of the Association for Machine Translation in the Americas (Volume 1: Research Track). 2024:250-260.

8. Suri G, Slater LR, Ziaee A, Nguyen M. Do large language models show decision heuristics similar to humans? A case study using GPT-3.5. J Exp Psychol Gen. 2024;153:1066-1075.
9. Cheung V, Maier M, Lieder F. Large language models show amplified cognitive biases in moral decision-making. P Natl Acad Sci USA. 2025;122:e2412015122.
10. Hagendorff T, Fabi S, Kosinski M. Human-like intuitive behavior and reasoning biases emerged in large language models but disappeared in ChatGPT. Nat Comput Sci. 2023;3:833-838.
11. McShane BB, Gal D, Duhachek A. Artificial intelligence and dichotomania. Judgm Decis Mak. 2025;20:e23.
12. Chen Y, Kirshner SN, Ovchinnikov A, Andiappan M, Jenkin T. A manager and an AI walk into a bar: does ChatGPT make biased decisions like we do? Manuf Serv Op. 2025;27:354-368.
13. Horowitz I, Plonsky O. LLM agents display human biases but exhibit distinct learning patterns. 2025. Available at https://arxiv.org/abs/2503.10248
14. Binz M, Schulz E. Using cognitive psychology to understand GPT-3. P Natl Acad Sci USA. 2023;120:e2218523120.
15. Kahneman D, Tversky A. Prospect theory: An analysis of decision under risk. Econometrica. 1979;47:363-391.
16. Robbins JM, Krueger JI. Social projection to ingroups and outgroups: A review and meta-analysis. Pers Soc Psychol Rev. 2005;9:32-47.
17. Krueger JI. Social projection as a source of cooperation. Curr Dir Psychol Sci. 2013;22:289-294.
18. Yechiam E, Hochman G. Losses as modulators of attention: Review and analysis of the unique effects of losses over gains. Psychol Bull. 2013;139:497-518.

19. Gal D, Rucker D. The loss of loss aversion: Will it loom larger than its gain? J Consum Psychol. 2018;28:497-516.
20. Yechiam E. Acceptable losses: The debatable origins of loss aversion. Psychol Res. 2019;83:1327-1339.
21. Rakow T, Cheung NY, Restelli C. Losing my loss aversion: The effects of current and past environment on the relative sensitivity to losses and gains. Psychon B Rev. 2020;27:1333-1340.
22. Brown AL, Imai T, Vieider F, Camerer CF. Meta-analysis of empirical estimates of loss-aversion. J Econ Lit. 2024;62:485-516.
23. Yechiam E, Zeif D. Loss aversion is not robust: A re-meta-analysis. J Econ Psychol. 2025;107:102801.
24. OpenAI. GPT-4 technical report. 2023. Available at arXiv:2303.08774.
25. Bai S, Chen K, Liu X, Wang J, Ge W, Song S, et al. Qwen2.5 technical report. 2024. Available at https://doi.org/10.48550/arXiv.2502.13923
26. DemandSage. ChatGPT statistics and facts (2025 update). Available at https://www.demandsage.com/chatgpt-statistics/
27. Cui Z, Li N, Zhou H. A large-scale replication of scenario-based experiments in psychology and management using large language models. Nat Comput Sci. 2025;5:627-634.
28. Hermann D. Determinants of financial loss aversion: The influence of prenatal androgen exposure (2D:4D). Pers Individ Differ. 2017;117:273-279.
29. Zeif D, Yechiam E. Loss aversion (simply) does not materialize for smaller losses. Judgm Decis Mak. 2022;17:1015-1042.
30. Walasek, L, Mullett TL, Stewart N. A meta-analysis of loss aversion in risky contexts. J Econ Psychol. 2024;103:102740.

31. Stella M, Hills TT, Kenett YN. Using cognitive psychology to understand GPT-like models needs to extend beyond human biases. P Natl Acad Sci USA. 2023;120:e2312911120.
32. Markman KD, Klein WMP, Suhr JA. Multiple explanation: A consider-an-alternative strategy for debiasing judgments. J Appl Soc Psychol. 2004;34:2272-2290.
33. Webb EC, Shu SB. Is broad bracketing always better? How broad decision framing leads to more optimal preferences over repeated gambles. Judgm Decis Mak. 2017;12:513-526.

# Supporting information

**S1 File. Supplementary information.** Additional experimental results, the debiasing experiment, loss aversion estimations using an orthodox approach, and the complete code of all studies.

Box 1: Study 1 and 2 decisions.

"Please let us know what would be your own preferences given:

a. A lottery providing 50% to win 10 Tenits and 50% to win 10 Tanas;

or alternatively get 10 Tenits for sure

b. A lottery providing 50% to win 10 Tenits and 50% to win 10 Tanas;

or alternatively get 10 Tanas for sure

c. A lottery providing 50% to win 10 Tenits and 50% to win 10 Tanas;

or alternatively get 8 Tenits for sure

d. A lottery providing 50% to win 10 Tenits and 50% to win 10 Tanas;

or alternatively get 8 Tanas for sure

e. A lottery providing 50% to win 10 Tenits and 50% to win 10 Tanas;

or alternatively get 5 Tenits for sure

f. A lottery providing 50% to win 10 Tenits and 50% to win 10 Tanas;

or alternatively get 5 Tanas for sure"

Box 2: Study 3 and 4 decisions.

"Lottery A: If the coin turns up heads, they lose $2; if the coin turns up tails, they win $6

Lottery B: If the coin turns up heads, they lose $3; if the coin turns up tails, they win $6

Lottery C: If the coin turns up heads, they lose $4; if the coin turns up tails, they win $6

Lottery D: If the coin turns up heads, they lose $5; if the coin turns up tails, they win $6

Lottery E: If the coin turns up heads, they lose $6; if the coin turns up tails, they win $6

Lottery F: If the coin turns up heads, they lose $7; if the coin turns up tails, they win $6"

<Study 4 only:>

Lottery G: If the coin turns up heads, they lose $8; if the coin turns up tails, they win $6"

Lottery H: If the coin turns up heads, they lose $9; if the coin turns up tails, they win $6"

Lottery I: If the coin turns up heads, they lose $9; if the coin turns up tails, they win $6"

"Please indicate for each of the following lotteries whether you would 'accept' that is play the lottery, for a chance of winning, or 'reject' it and not receive anything (you can't be 'neutral')."

Figure 1: Study 1 and 2 results. Proportions of accepted safe options by ChatGPT and Qwen, when denominated in a (human) chosen currency, given different expected values (EVs). In Study 1 the chosen currency by humans is the preferred one while in Study 2 there is no logical evidence of that. The differences between the preferred/ unpreferred columns denote LLM biases due to subjective preference.

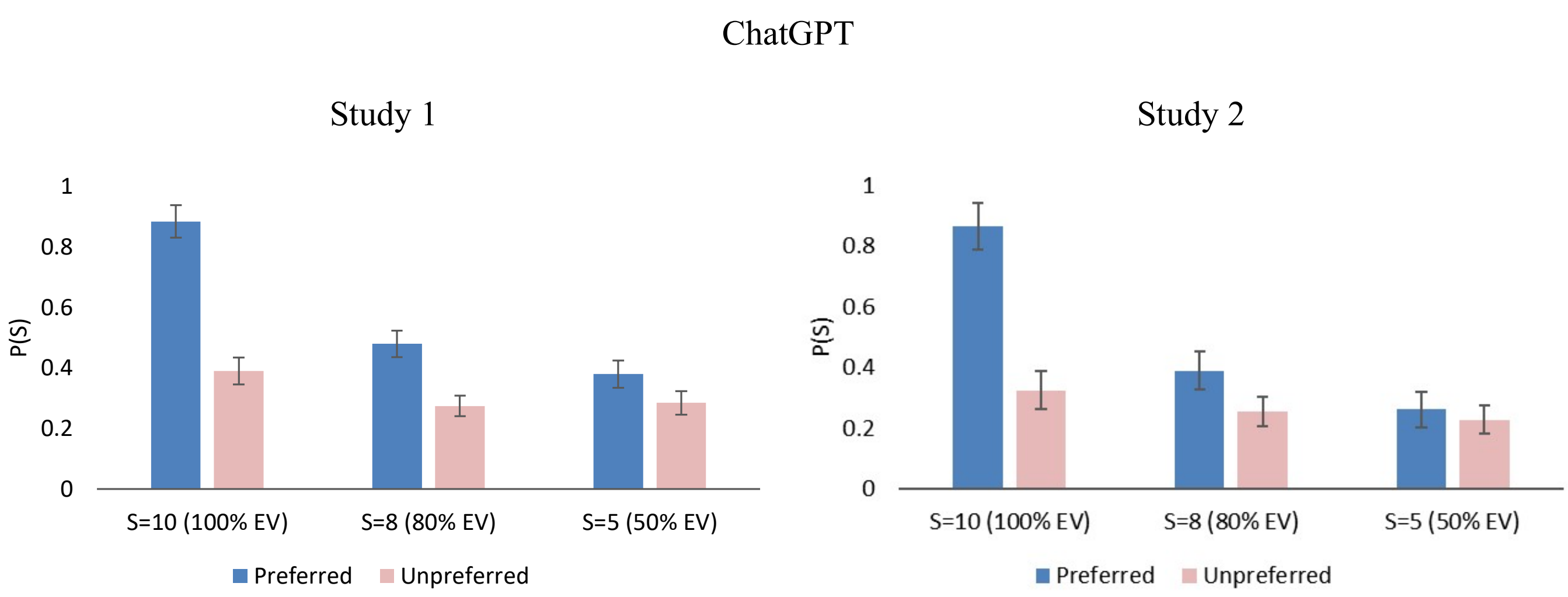

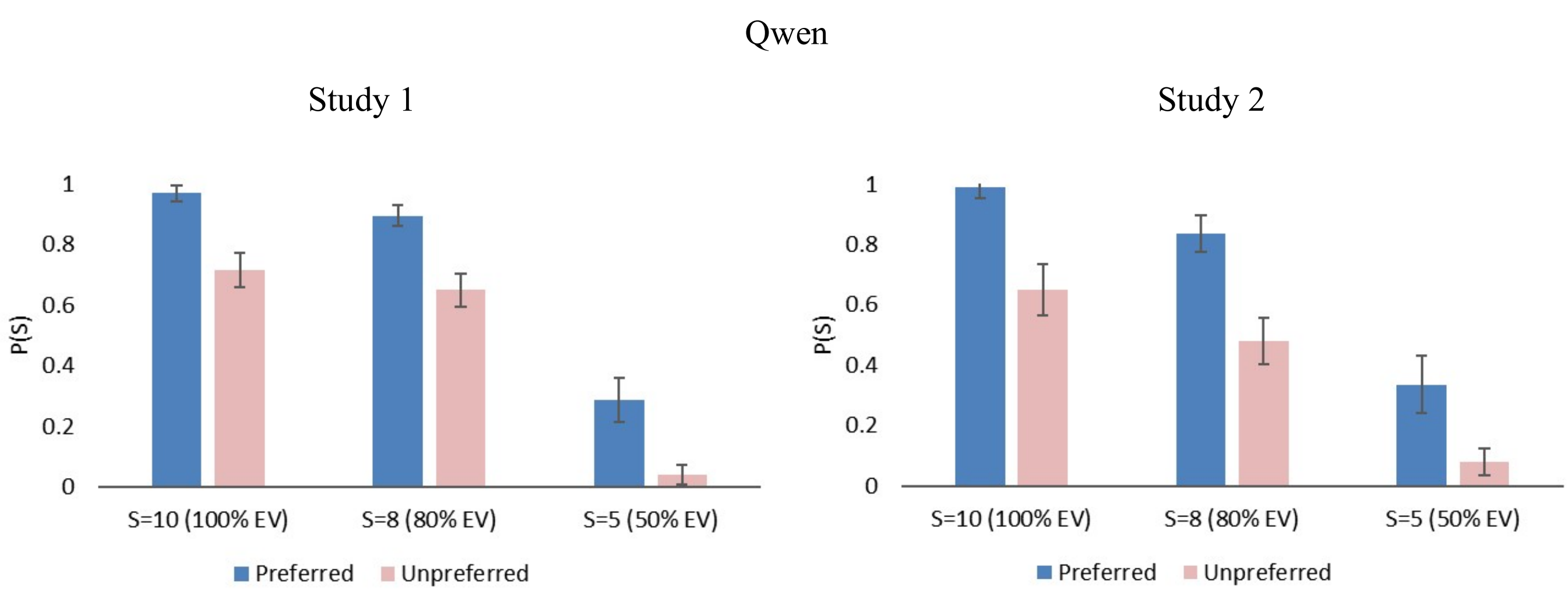

Figure 2: Study 3 results. Proportions of accepted lotteries involving equiprobable gains and losses (shown on the abscissa) by ChatGPT and Qwen, when presented with experimental results from a large/small sample of loss averse/gain seeking humans.

ChatGPT

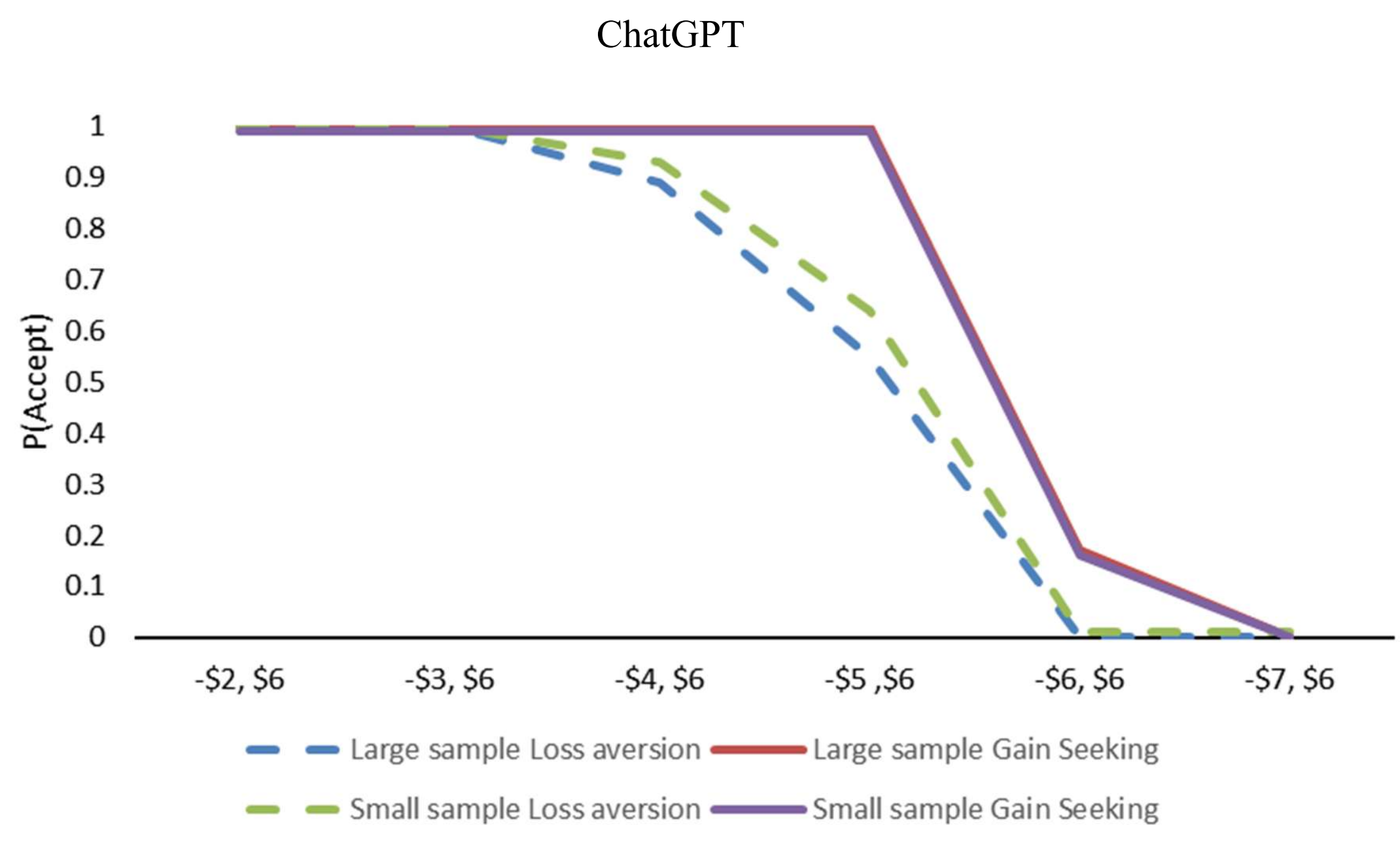

Qwen

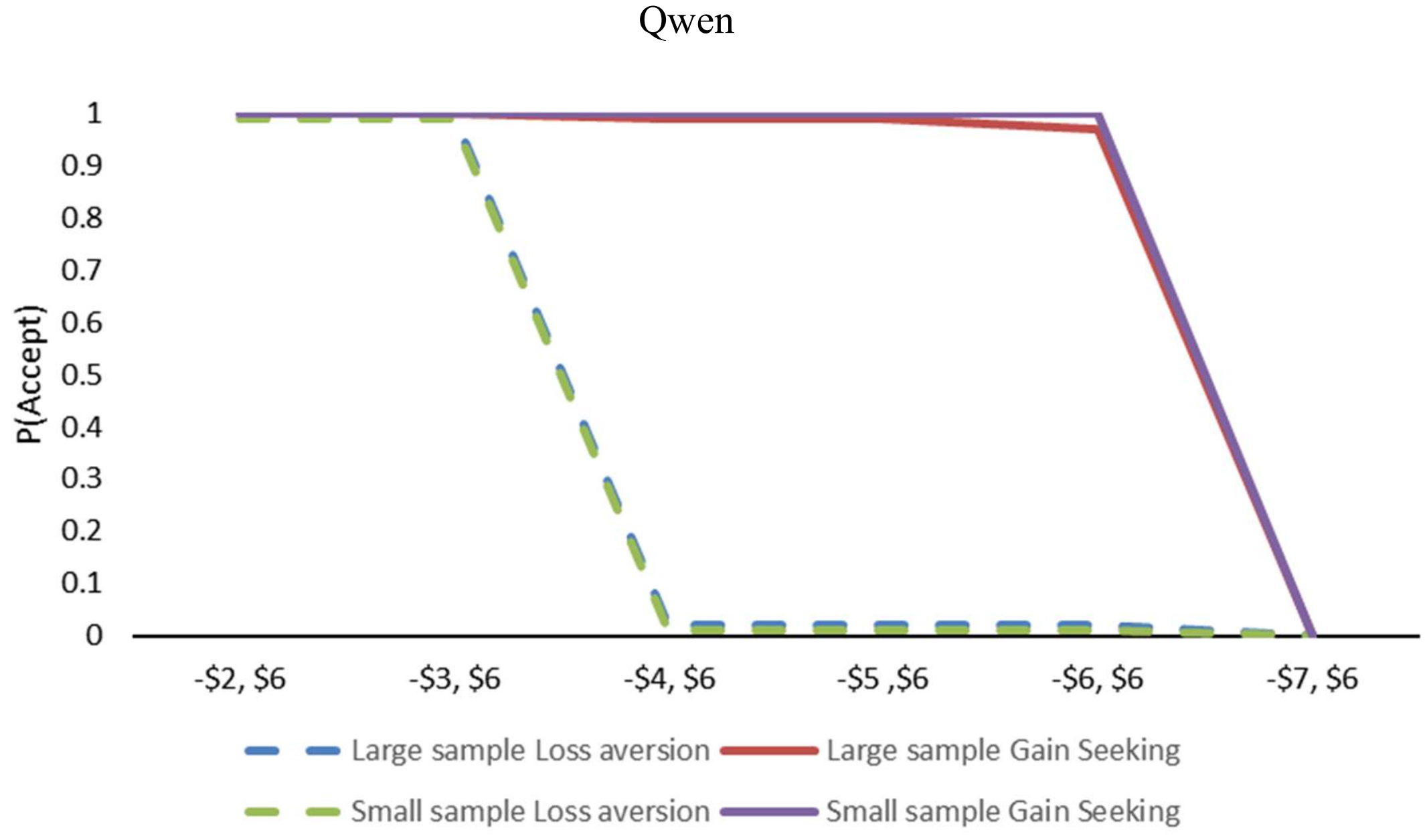

Figure 3: Study 4a (top two panels) and 4b (bottom panel) results. Left: Proportions of accepted lotteries involving equiprobable gains and losses (shown on the abscissa) by LLMs, when basing the answer on different scientific papers concerning loss aversion. Right: Mean estimated loss aversion parameter in the original human study (wireframe columns) and in LLMs' decisions based on the study.

ChatGPT

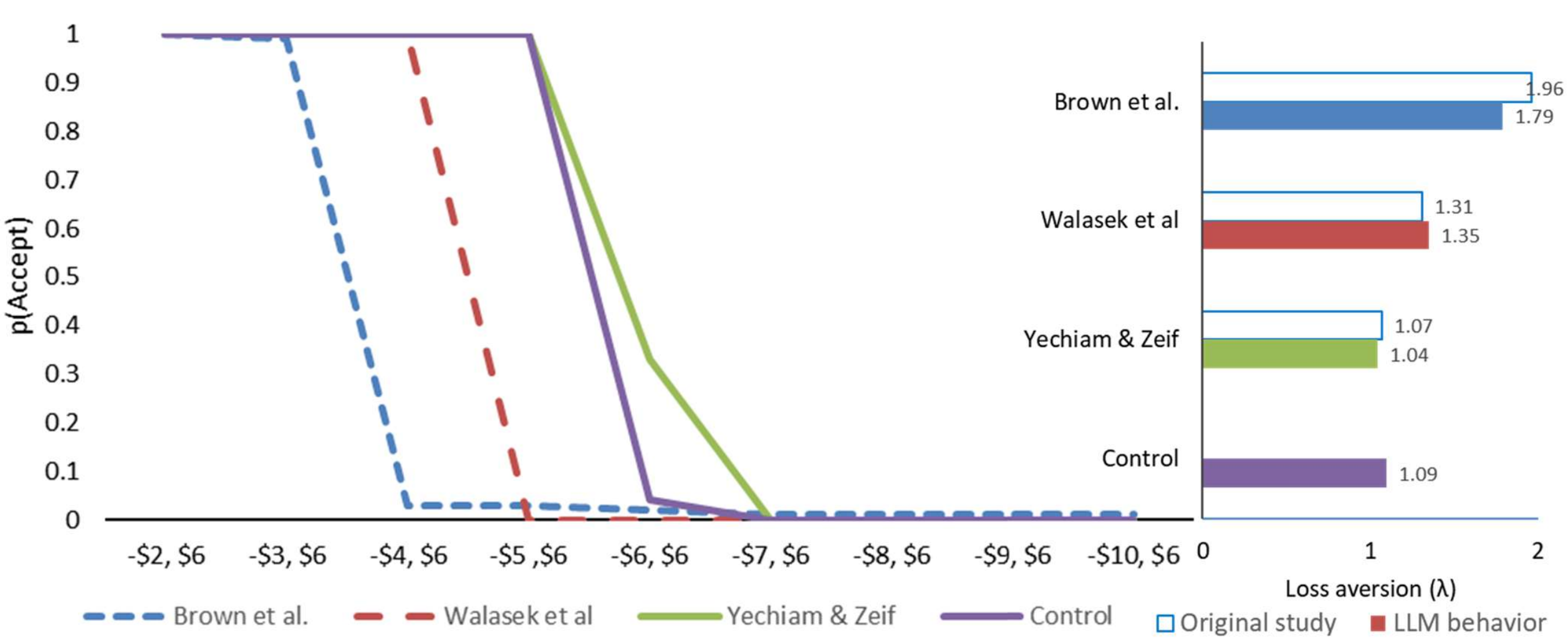

Qwen

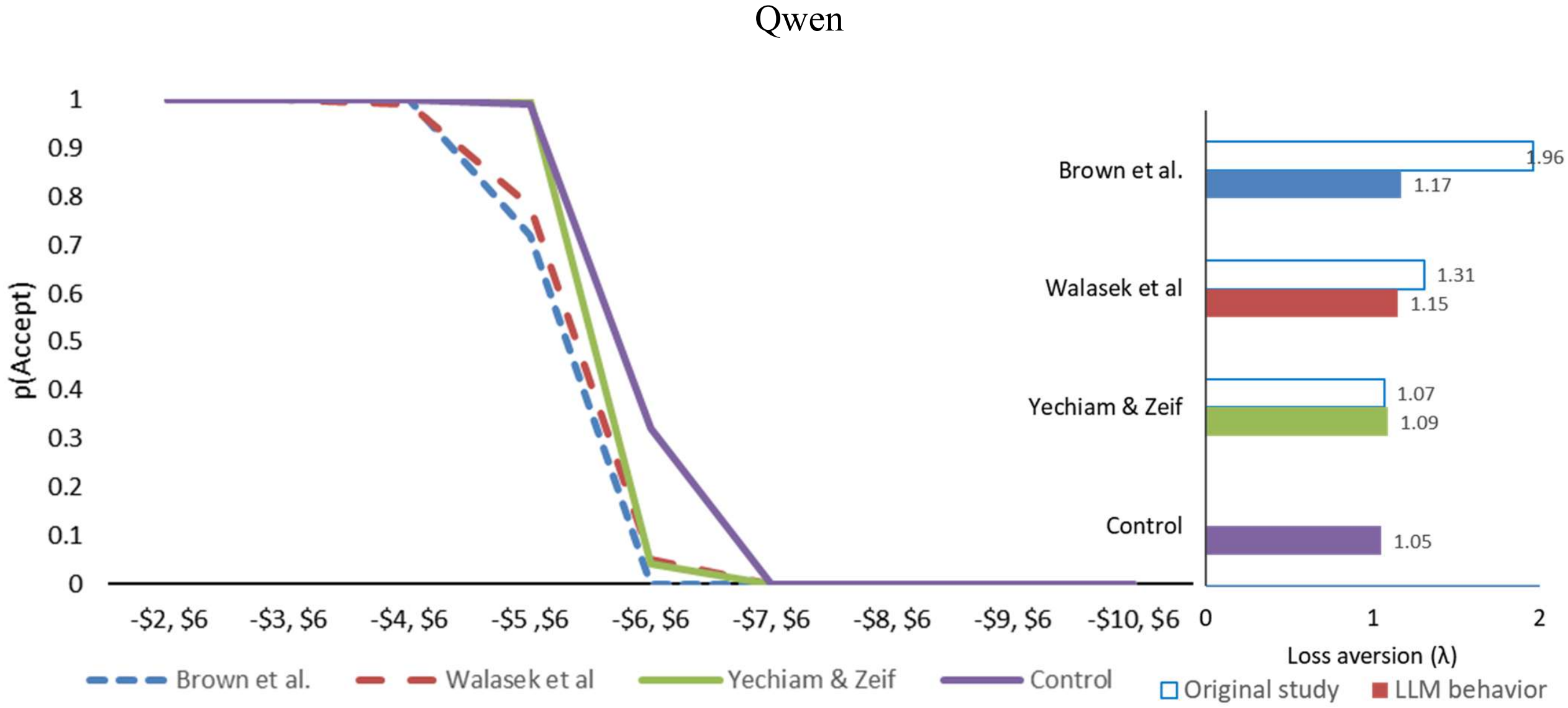

ChatGPT: Full text of scientific papers

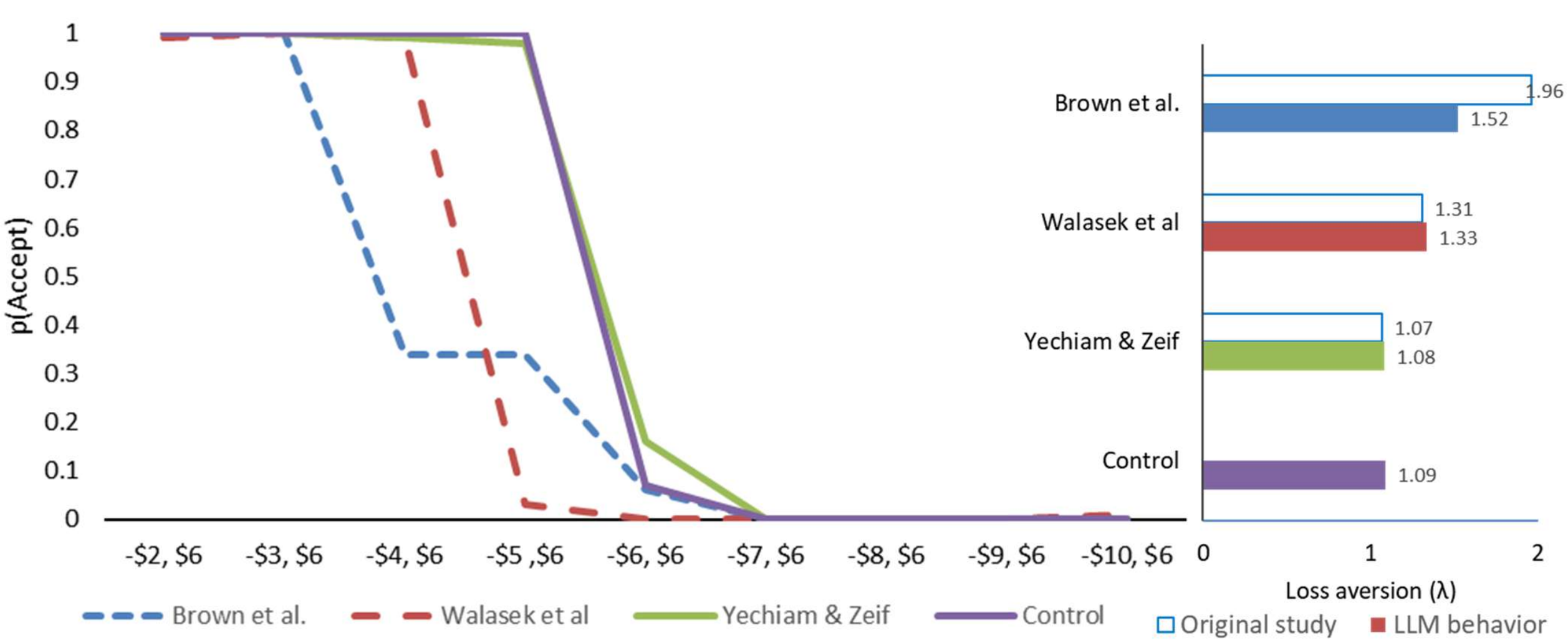